\documentclass[a4paper]{spie}  

\usepackage{amsmath,amsfonts,amssymb}
\usepackage{graphicx}
\usepackage[colorlinks=true, allcolors=blue]{hyperref}
\usepackage{todonotes}
\usepackage{subcaption}
\usepackage[nolist]{acronym}
\usepackage{multirow}

\title{Weakly Supervised Airway Orifice Segmentation in Video Bronchoscopy}

\author[a]{Ron Keuth}
\author[a]{Mattias Heinrich}
\author[b]{Martin Eichenlaub}
\author[a]{Marian Himstedt}

\affil[a]{Institute of Medical Informatics, University of Lübeck, Germany}
\affil[b]{University Heart Center Freiburg-Bad Krozingen, Germany}

\authorinfo{Further author information: E-mail: ron.keuth@student.uni-luebeck.de, marian.himstedt@uni-luebeck.de}

\begin{document} 
\begin{acronym}
    \acro{cnn}[CNN]{Convolutional Neural Network}
    \acro{lraspp}[Lite R-ASPP]{Lite Reduced Atrous Spatial Pyramid Pooling}
    \acro{cvc}[CVC-DS]{Dataset from the Computer Vision Center's interactive and augmented modelling group}
    \acro{rb}[RB-DS]{Real Bronchoscopy Dataset}
    \acro{dsc}[DSC]{Dice Similarity Score}
    \acro{gan}[GAN]{Generative Adversarial Networks}
    \acro{amcd}[AMCD]{Average Minimum Centriod Distance}
    \acro{vb}[VB]{Video Bronchoscopy}
    \acro{copd}[COPD]{Chronic Obstructive Pulmonary Disease}
    \acro{icu}[ICU]{Intensive Care Units}
    \acro{emt}[EMT]{Electromagnetic Tracking}
\end{acronym}

\maketitle

\section{Description of purpose}\label{sec:intro}
\ac{vb} is commonly applied in conjunction with lung diseases. It is a fundamental procedure for diagnosis of lung cancer enabling biopsy of deep airway tissue. In addition to that, \ac{vb} is routinely conducted for monitoring \ac{copd} patients and clarification of acute respiratory problems at \ac{icu}. The navigation within the bronchial tree is challenging and physically demanding for physicians due to homogenous textures and perceptually similar appearance of bronchial orifices. This is particularly the case in the absence of prior CT scans and \ac{emt} systems at \ac{icu}s. Airway orifice segmentation which is the main objective of this paper enables image-based guidance, e.g. by providing graphical overlays on top of the \ac{vb} images. In conjunction with \ac{emt} or image-based tracking\cite{wang_visual_2020} these overlays can be accompanied by airway labels w.r.t. generic or interpatient lung models which however is not addressed by the approach presented in this paper. The variety of tissue appearance, illumination, image artifacts, secrete and patient anatomy poses a challenge for airway segmentation which we aim to address by deep learning-based approaches. However, this is currently hampered due to a lack of readily available ground truth labels motivating the incorporation of traditional (non-learning-based) methods as weak supervision. In particular, we incorporate an airway phantom dataset collection accompanied by ground truth depth images to generate airway orifice labels for training a deep learning-based segmentation model. 
Our proposed methods are also developed with a focus on their complexity and runtime, keeping them real-time capable even on low-end devices for intervention guidance.

\section{Methods}
\subsection{Datasets}
We utilize three datasets for training, validation and testing our method: The \textbf{Phantom} dataset \cite{visentini-scarzanella_deep_2017} consisting of about 30k RGB and depth images captured within a simplified model. For in-vivo evaluation, we use 125 samples of 20 different bronchoscopies with their annotated segmentations from the public \textbf{CVC-DS} dataset \cite{sanchez_-line_2014} and about 100 frames of the private dataset \textbf{RB-DS} with expert annotations.

\begin{figure}
    \centering
    
    \begin{subfigure}[t]{.25\textwidth}
        \centering
        \includegraphics[width=\linewidth, trim={20cm, 7cm, 20cm, 9cm},clip]{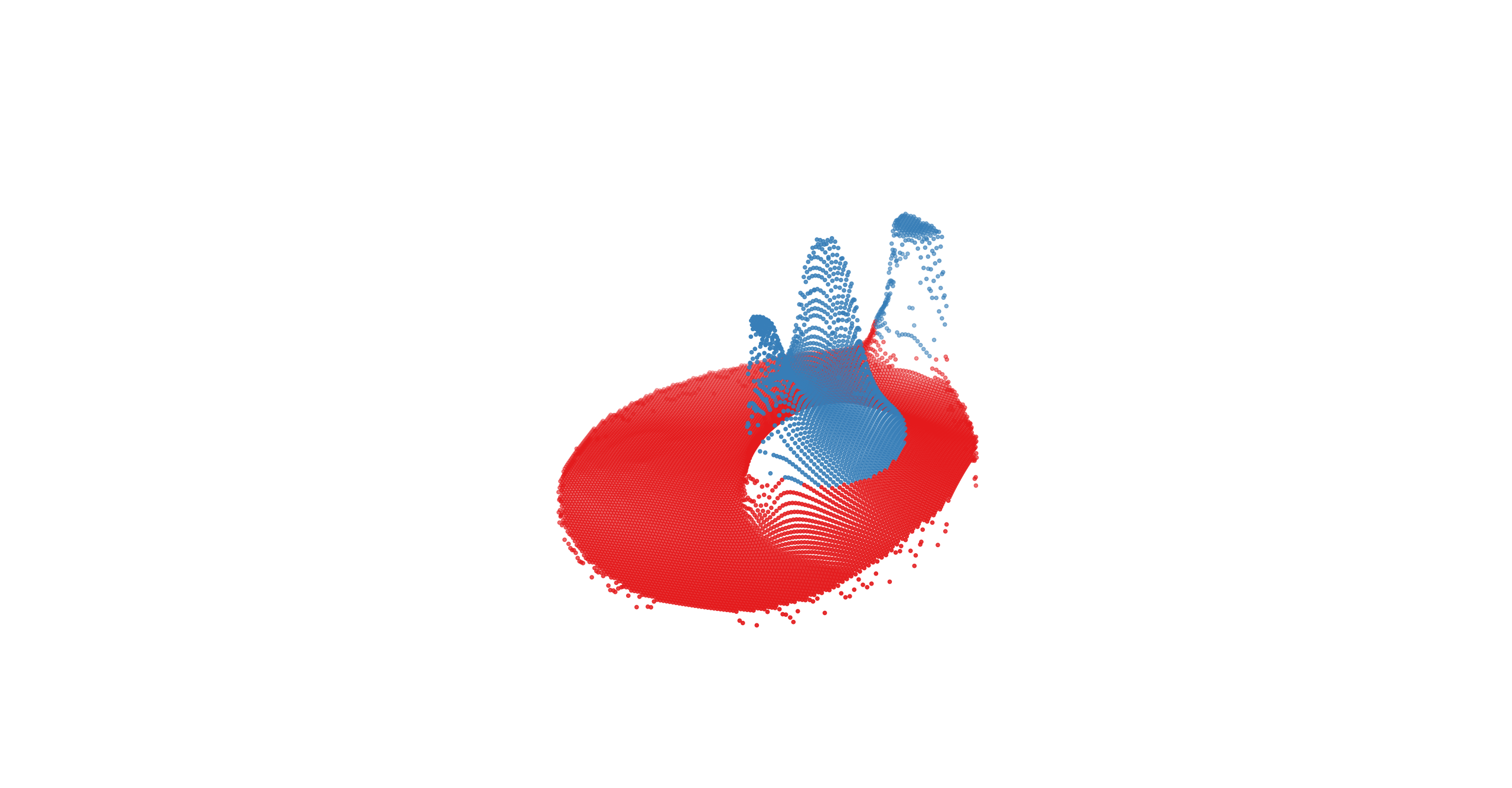}
        \caption{Detecting airways using $k$-means ($k=2$) on the depth data.}
        \label{fig:k_means_pc}
    \end{subfigure}
    \hspace{4cm}
    \begin{subfigure}[t]{.25\textwidth}
        \centering
        \includegraphics[width=\linewidth, trim={20cm, 6.2cm, 20cm, 10cm},clip]{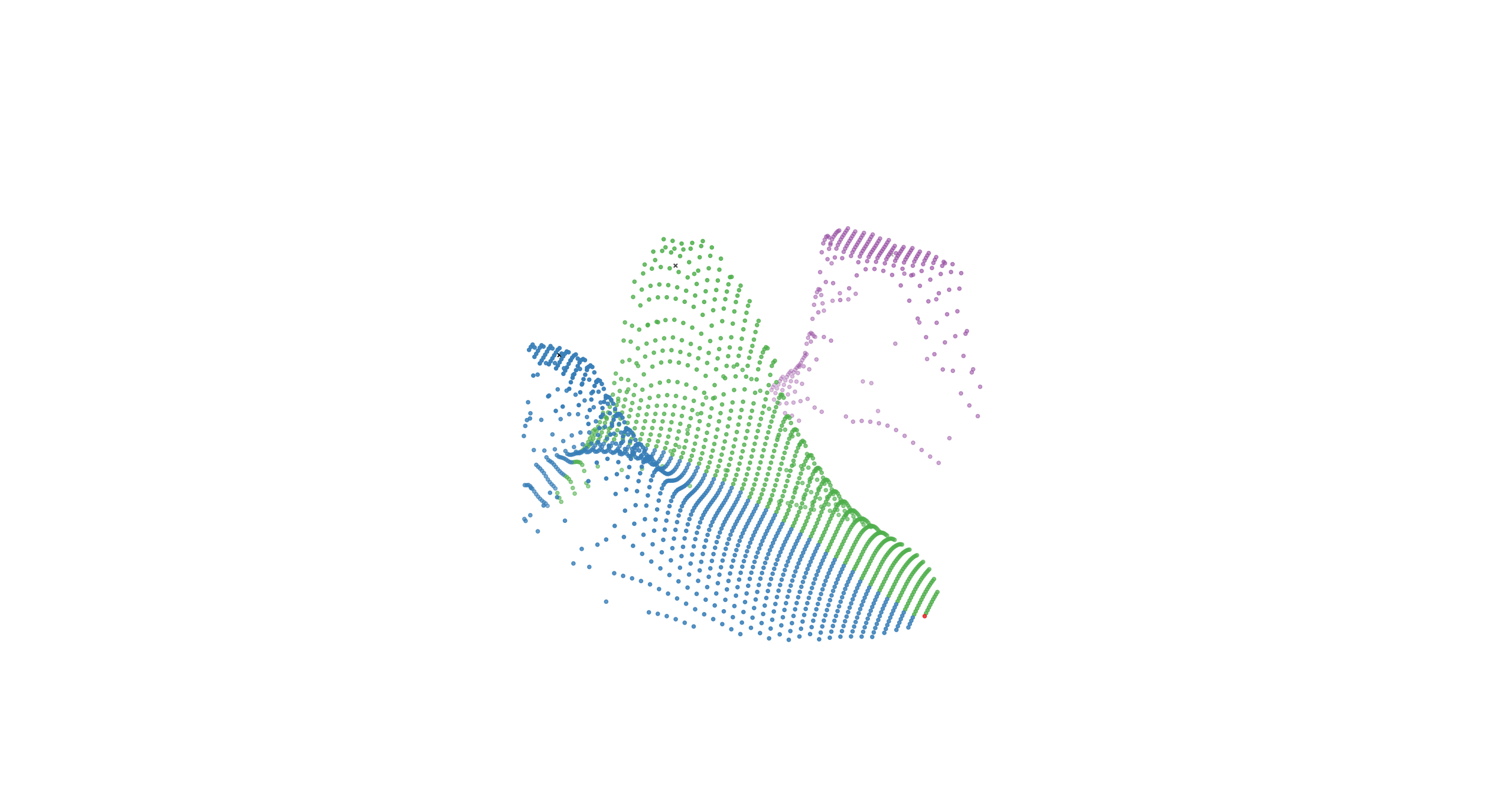}
        \caption{Extracting the individually airways with a compact marker-based watershed.}
        \label{fig:watershed_pc}
    \end{subfigure}
    
    \medskip
    
    \begin{subfigure}[t]{.2\textwidth}
        \centering
        \includegraphics[width=\linewidth]{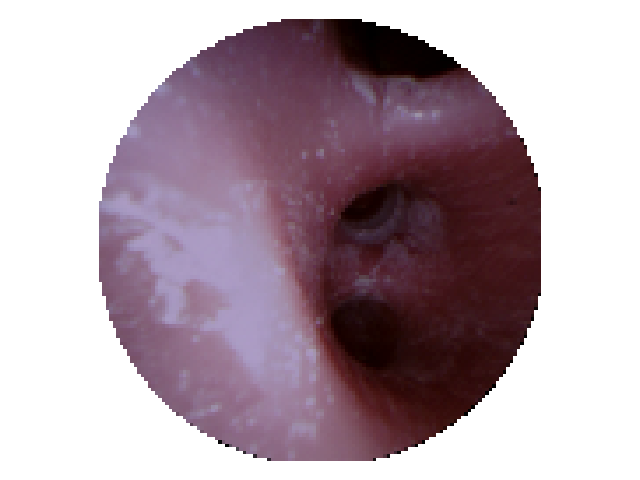}
        \caption{Corresponding RGB bronchoscopy image.}
        \label{fig:rgb_img}
    \end{subfigure}
    \hspace{.5cm}
    \begin{subfigure}[t]{.2\textwidth}
        \centering
        \includegraphics[width=\linewidth]{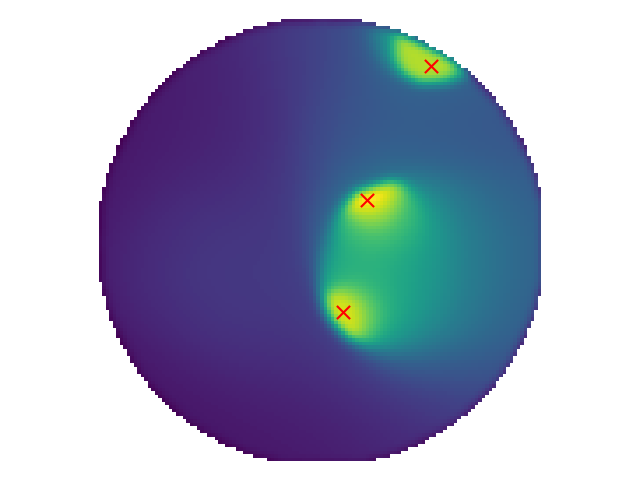}
        \caption{Smoothed depth image with its local peaks (red markers).}
        \label{fig:depth_img_with_markers}
    \end{subfigure}
    \hspace{.5cm}
    \begin{subfigure}[t]{.2\textwidth}
        \centering
        \includegraphics[width=\linewidth]{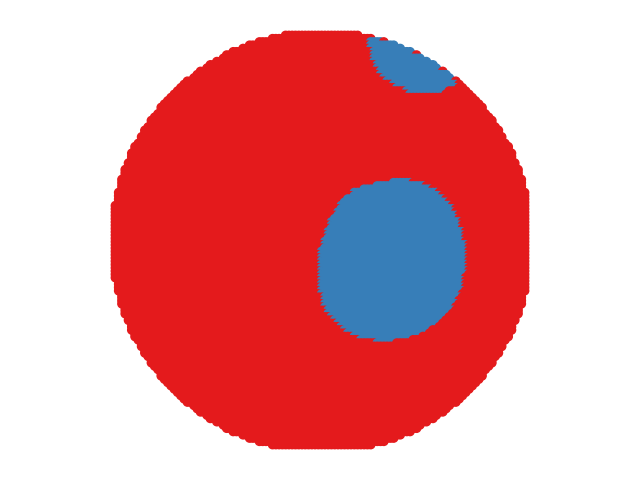}
        \caption{Obtained binary segmentation map.}
        \label{fig:binary_lbl}
    \end{subfigure}
    \hspace{.5cm}
    \begin{subfigure}[t]{.2\textwidth}
        \centering
        \includegraphics[width=\linewidth]{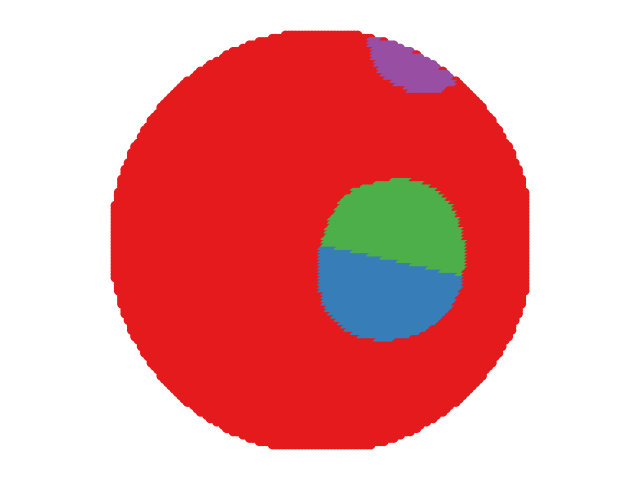}
        \caption{Obtained airway instances segmentation map.}
        \label{fig:multi_lbl}
    \end{subfigure}
    
    \caption{The stages of our segmentation pipeline; A $k$-means determine the global airway labels, followed by compact wahtershed, which use the locale peaks from the smoothed depth image as markers to discriminate the individual airway instances.}
    \label{fig:seg_pipeline}
\end{figure}

\subsection{Data-Driven Methods for Instance Orifice Segmentation}
Fig.\ref{fig:seg_pipeline} shows the different steps of our proposed pipeline to generate an instance orifice segmentation map from a given depth image. A $k$-means determines the two classes ($k=2$) of airway and other tissue, considering only the depth distribution (see Fig.\ref{fig:k_means_pc}). The obtained global airway labels are then used to generate a binary segmentation map (see Fig.\ref{fig:binary_lbl}) and to define the region of interest for the next stage. For determining different airway instances, we low-pass filter the depth image using an efficient box filter (3$\times$average pooling with kernel size=3) followed by a non-maximum suppression, where the peaks have to be $5\%$ of the image resolution apart from each other (see Fig.\ref{fig:depth_img_with_markers}). These peaks define the markers for the compactness marker-based watershed algorithm\cite{protzel_compact_2014}, which runs on the inverted depth image as input. The watershed perfectly models the nature of the instance airway segmentation problem, allowing different depth values of adjacent airways and let their segmentation flow smoothly into each other (see Fig.\ref{fig:watershed_pc}). The result of the watershed is finally composed together with the global labels to an instance orifice segmentation map (see Fig.\ref{fig:multi_lbl}).

The pipeline is very real-time capable, running with roughly $130\,\text{Hz}$ on a laptop CPU\footnote{INTEL i5-7200U 2C/4T@3.1GHz}.
One current limitation of our pipeline is the segmentation of at least one orifice, even if it is a false positive. However, because such false-positive leans to cover an unusually huge area, the problem can be solved by defining a heuristic such as a relative threshold over the area covered by one orifice instance.

\subsection{\acs{cnn} Architecture}
For this paper, we solely focus on the binary orifice segmentation to enable the use of a \ac{lraspp}\cite{howard_searching_nodate} as an efficient \ac{cnn} architecture for segmentation. We use an encoder pretrained on ImageNet for training. The high texture and illumination variety from the synthetic to the in-vivo \acp{vb} introduces a domain gap. To narrow this gap, we apply multiple data augmentation methods. Therefore we follow a guideline\cite{cubuk_autoaugment_2019} for an intensity value augmentation of the RGB images, randomly choosing transformations like color jitter, quantization and histogram equalization. 
In in-vivo \ac{vb} the operator rotate the endoscope a lot during navigation, making rotation invariance a real-world requirement for our methods. We achieve that by rotating the images with radiants randomly sampled from $[0, 2\pi]$\cite{yoo_deep_123}. To prevent the model's parameter from becoming too complex and overfitting the synthetic data, we use weight decay and its AdamW implementation\cite{loshchilov_decoupled_2019} during training.


\subsection{Metric for Evaluation}\label{subsec:metric}
A well established metric to evaluate the overlapping of two segmentations is the \acf{dsc}. However, the \ac{dsc} alone has only limited significance in our context. This is because an airway's orifice has, unlike e.g. a liver, no clear organ boundaries, resulting in a high inter and also intra observer variability in the in-vivo segmentation ground truths. The first column of Fig.\ref{fig:pred_cvc} shows a good example for such a situation, where four different sized segmentations are provided for the same airway orifice with all being correct, but resulting in an underestimated \ac{dsc}.

As a solution, we also consider the distance of the first moments (centers of gravity) of the individual airway orifice instance segmentations. Moments are scale invariant and therefore well-suited for this use case. We convert the distances of the first moments into our \ac{amcd} metric as followed: Having $N$ airways with their ground truth segmentation and $M$ predicted segmentation, we calculate their first moments in $C\in \mathbb{R}^{N\times2}$ and $\hat{C}\in\mathbb{R}^{M\times2}$ respectively. The minimal distance of the center $d_{c_i}$ is then the minimum of the euclidean distance:
\begin{equation}\label{eq:d_c}
    d_{c_i} = \underset{\hat{c}_j}{\arg\min}||c_i-\hat{c}_j||_2
\end{equation}
with $i\in N$ and $j\in M$. We finally obtain the \ac{amcd} for the overall image by the mean $\bar{d}_c=\frac{1}{N}\sum_{i\in N}d_{c_i}$.
However, we decided to include the \ac{dsc} due to its scientific importance even though its signficance is rather limited for the evaluation of our approach.

\subsection{Trainings and Evaluation}
We train a \ac{lraspp} instance on the training split of the phantom dataset and evaluate its performance on the test split and also on the two in-vivo \ac{vb} datasets. For each of the two in-vivo datasets a model is also trained to examine if the semantic airway knowledge gained by the synthetic data is comparable to the one by the in-vivo data. It has to be mentioned that we decided against a cross validation on the in-vivo datasets due to their limited sizes. Thus, the performances on their own dataset do not demonstrate their ability to generalize to unseen data.

\section{Results}
\begin{table}[]
    \centering
    \caption{Quantitative results. We use the \acf{dsc} and the \acl{amcd} $\bar{d}_c$ (see Eq.\ref{eq:d_c}) within the image resolution of $128^2$. \textbf{Please remind the limited significance of the \ac{dsc} in our context (see Sec.\ref{subsec:metric}).}}
    \begin{tabular}{|l|l|c|c|c|}\hline
    train dataset & metric & \multicolumn{3}{|c|}{test datasets}\\\hline
    \multicolumn{2}{|c|}{}&Phantom[test]&\acs{cvc}&\acs{rb}\\
    \hline
    \multirow{2}{*}{Phantom[train]} & \acs{dsc} & $73.48\pm17.72$ & $60.91\pm17.83$ & $50.58\pm16$ \\
                              & $\bar{d}_c[\text{px}]$ & $20.41\pm11.55$ & $8.41\pm7.25$   & $14.58\pm10.75$\\\hline
    \multirow{2}{*}{\acs{cvc}}       & \acs{dsc} & $36.67\pm21.35$ & $85.79\pm7.27$ & $61.51\pm22.03$ \\
                              & $\bar{d}_c[\text{px}]$ & $20.19\pm17.52$ & $2.76\pm2.41$   & $9.5\pm11.27$\\\hline
    \multirow{2}{*}{\acs{rb}}        & \acs{dsc} & $32.72\pm12.66$ & $80.95\pm9.66$ & $80.57\pm0.1$ \\
                              & $\bar{d}_c[\text{px}]$ & $16.74\pm9.44$ & $3.55\pm3.43$   & $4.37\pm3.9$\\\hline
    \end{tabular}
    \label{tab:metric_scores}
\end{table}

\begin{figure}
    \centering
    \begin{subfigure}[t]{.32\linewidth}
        \centering
        \includegraphics[width=\linewidth]{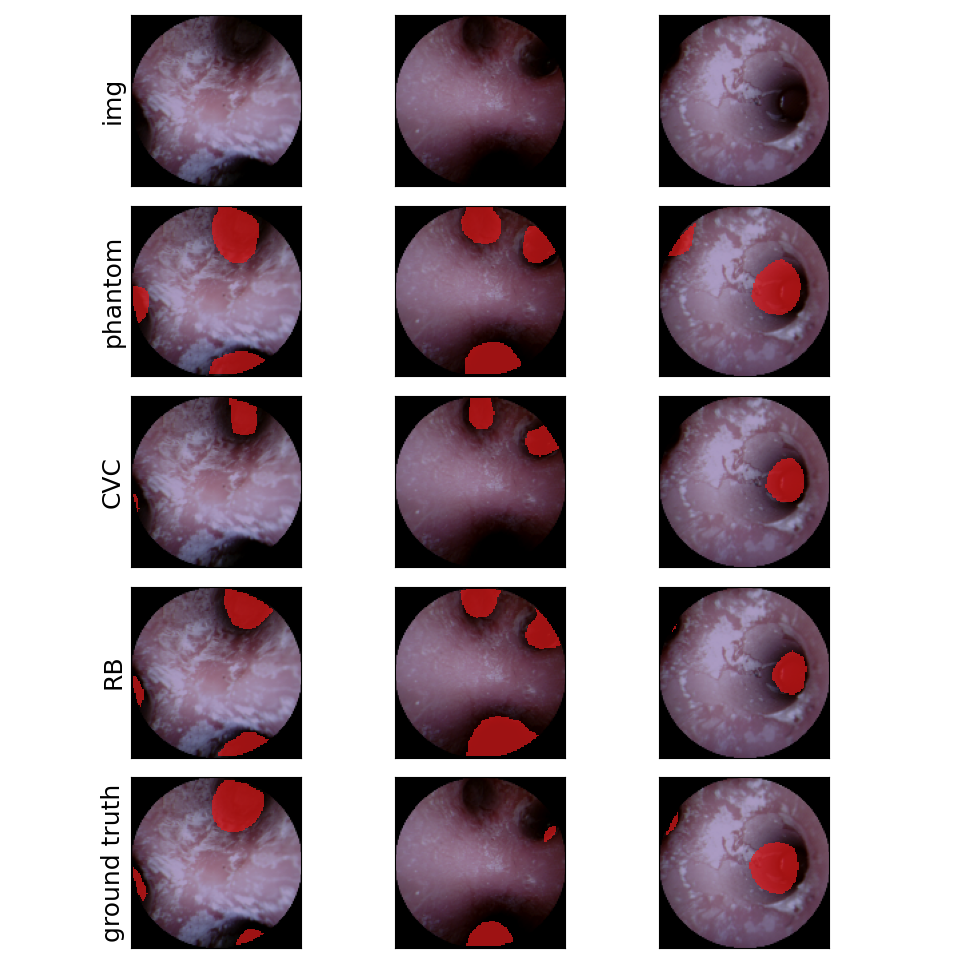}
        \caption{Test split of the phantom dataset}
        \label{fig:pred_phantom}
    \end{subfigure}
    \begin{subfigure}[t]{.32\linewidth}
        \centering
        \includegraphics[width=\linewidth]{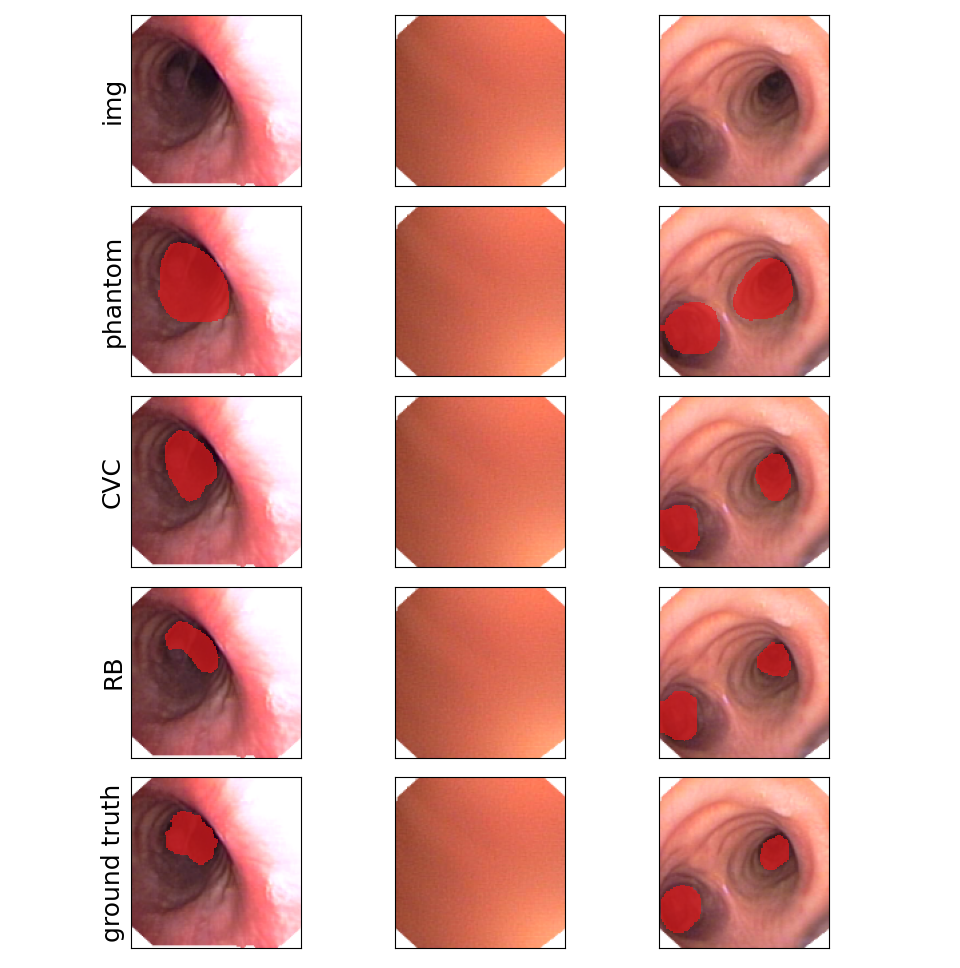}
        \caption{in-vivo \acs{cvc}}
        \label{fig:pred_cvc}
    \end{subfigure}
    \begin{subfigure}[t]{.32\linewidth}
        \centering
        \includegraphics[width=\linewidth]{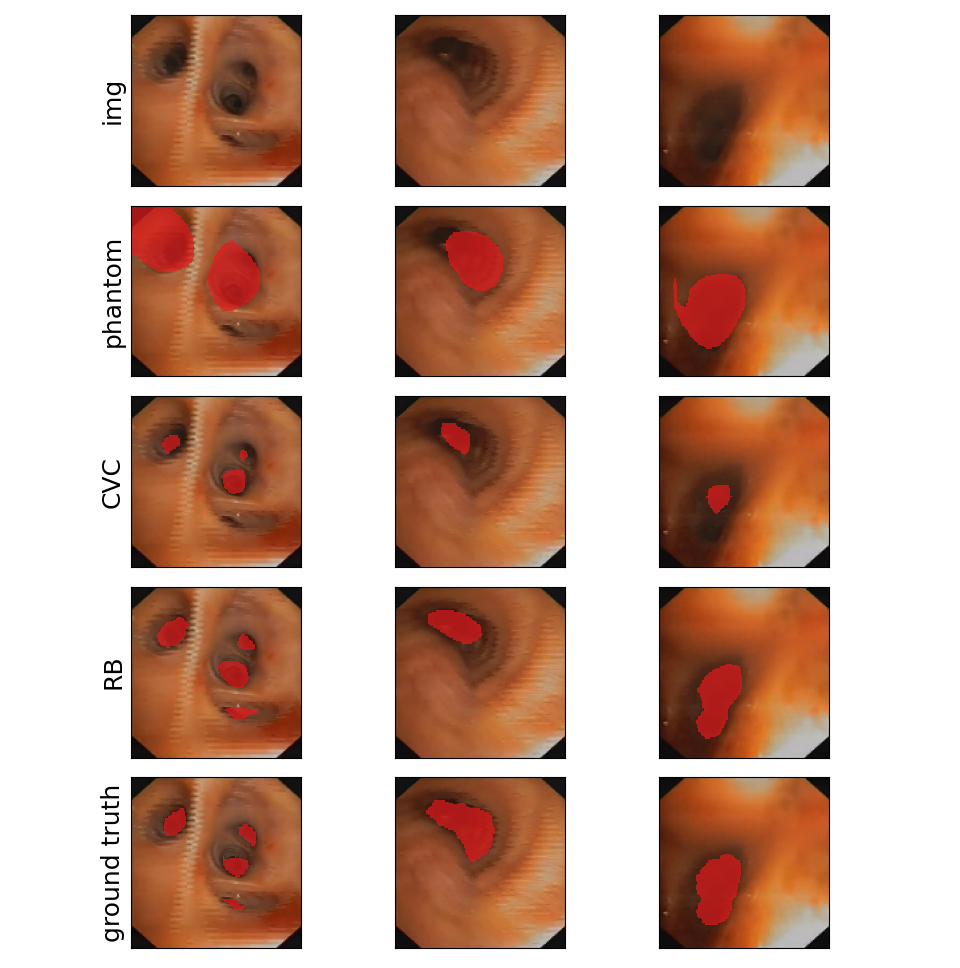}
        \caption{in-vivo \acs{rb}}
    \end{subfigure}
    \caption{Qualitative Results. The first row shows the input image, the last one its corresponding ground truth. The other row labels describe the model's training dataset and the figure's caption the dataset of the images.}
    \label{fig:seg_plot}
\end{figure}

The qualitative results in Fig.\ref{fig:seg_plot} demonstrate that the model trained on the phantom dataset was able to extract a semantic knowledge about airways from the noisy ground truth generated based on depth images using our proposed data-driven pipeline. In some cases, it even outperforms this noisy ground truth, particularly when airways were not detected properly due to their low depth profiles (see Fig.\ref{fig:pred_phantom}). This happens if an orifice belongs to an airway, that has a high angle to the camera perspective, and therefore only its wall remains visible.
The data augmentation method \cite{cubuk_autoaugment_2019} was able to close the domain gap from the phantom to the in-vivo datasets, when comparing the model's performance to the models being directly trained on the in-vivo data. Even cases without any airways present were correctly predicted, even though such situations were not directly covered by the ground truth during training (see Fig.\ref{fig:pred_cvc}). Having those results in mind, our training can be considered as successful for the real world use case.

Our quantitative results shown in Tab.\ref{tab:metric_scores} have only limited significance for the Phantom dataset due to lack of real ground truth, which is compensated by our noisy generated one. However, the ground truth by the in-vivo dataset was created by human experts and is, beside the very high observer variability, reliable. The model trained on the Phantom dataset shows a great domain robustness, considering the resulting \ac{dsc}. The domain gap seems larger coming from the in-vivo domain on the first sight. But this is a perfect example for our \ac{amcd} $\bar{d}_c$ and its motivation; even if the \ac{dsc} on the phantom data is twice as low as with the model trained on the phantom dataset, the \ac{amcd} is in the very same range, going hand in hand with the correct prediction visible in the qualitative results. This is due to the observer variability of the ground truth segmentation, resulting in different diameters of the segmentations depending on the observer.

Considering the quantitative but especially the qualitative results, the training on the Phantom data with the noisy ground truth as weak supervision enables the learning of a semantic knowledge of airways. We had also shown that this knowledge can be transferred to in-vivo data while reaching comparable performance as the models directly trained on these datasets. 

\section{New or breakthrough work to be presented}
This paper presents a novel approach for weakly-supervised CNN-based airway orifice segmentation in video bronchoscopy which is trained on phantom and evaluated on in-vivo patient data. To our best knowledge, this is the first paper presenting a deep learning-based approach omitting the use of depth images for inference. 

\section{Conclusion}
In this work, we presented a very real-time capable pipeline that extracts airway segmentations from a given bronchoscopy depth image using efficient data-driven classical methods. However, this method has some disadvantages: On the one hand it requires a depth image, which is not native given by the endoscope due to hardware limitation and therefore has to be generated via a complex non-linear domain translation like a \ac{gan}. On the other hand, a data-driven approach is not equal to a semantic understanding. Considering this and due to the lack of robustness to some edge cases, we consider this pipeline alone as not suitable for real world applications.
However, paired with the RGB image of the bronchoscopy, these generated segmentation maps can be used as weak supervision during training of a shallow \ac{cnn} for airway orifice segmentation. We showed that this model being trained on phantom data gains tremendous semantic knowledge of airway structures overcoming noisy ground truth on edge cases and is even applicable directly to in-vivo \ac{vb} due thanks to a substantial data augmentation.
With all of this, our proposed method allows the generation of segmentation masks directly on RGB images without the need of hand annotated datasets. We argue that this direct prediction from RGB images is superior to the segmentation approaches on the depth images, because it comes without the risks of a domain translation from RGB to depth via \acp{gan}, which is mainly based on unsupervised manner of the \ac{gan} training, likely to cause the generation of wrong anatomies\cite{shin_abnormal_2018} like additional or absent airway branches in the synthesized depth images.


\bibliography{report} 

\begin{thebibliography}{1}

\bibitem{wang_visual_2020}
Wang, C., Oda, M., Hayashi, Y., Villard, B., Kitasaka, T., Takabatake, H.,
  Mori, M., Honma, H., Natori, H., and Mori, K., ``A visual {SLAM}-based
  bronchoscope tracking scheme for bronchoscopic navigation,'' {\em
  International Journal of Computer Assisted Radiology and Surgery}~{\bf 15},
  1619--1630 (Oct. 2020).

\bibitem{visentini-scarzanella_deep_2017}
Visentini-Scarzanella, M., Sugiura, T., Kaneko, T., and Koto, S., ``Deep
  monocular {3D} reconstruction for assisted navigation in bronchoscopy,'' {\em
  International Journal of Computer Assisted Radiology and Surgery}~{\bf 12},
  1089--1099 (July 2017).
\newblock Publisher: Springer Verlag.

\bibitem{sanchez_-line_2014}
Sánchez, C., Bernal, J., Gil, D., and Sánchez, F.~J., ``On-{Line} {Lumen}
  {Centre} {Detection} in {Gastrointestinal} and {Respiratory} {Endoscopy},''
  in [{\em Clinical {Image}-{Based} {Procedures}. {Translational} {Research} in
  {Medical} {Imaging}}{\nolinebreak\hspace{0.1em}]},  Erdt, M., Linguraru,
  M.~G., Oyarzun~Laura, C., Shekhar, R., Wesarg, S., González~Ballester,
  M.~A., and Drechsler, K., eds.,  31--38, Springer International Publishing,
  Cham (2014).

\bibitem{protzel_compact_2014}
Protzel, P., ``Compact {Watershed} and {Preemptive} {SLIC}: {On} improving
  trade-offs of superpixel segmentation algorithms,'' (2014).

\bibitem{howard_searching_nodate}
Howard, A., Wang, W., Chu, G., Chen, L.-c., Chen, B., and Tan, M., ``Searching
  for {MobileNetV3},''
\newblock arXiv: 1905.02244v5.

\bibitem{cubuk_autoaugment_2019}
Cubuk, E.~D., Zoph, B., Mane, D., Vasudevan, V., and Le, Q.~V., ``Autoaugment:
  {Learning} augmentation strategies from data,'' {\em Proceedings of the IEEE
  Computer Society Conference on Computer Vision and Pattern Recognition}~{\bf
  2019},  113--123 (June 2019).
\newblock arXiv: 1805.09501v3 ISBN: 9781728132938.

\bibitem{yoo_deep_123}
Yoo, J.~Y., Kang, Y., Park, J.~S., Cho, Y.-J., Park, S.~Y., Yoon, H.~I., Park,
  S.~J., Jeong, H.-G., and Kim, T., ``Deep learning for anatomical
  interpretation of video bronchoscopy images,'' {\em Scientific Reports
  {\textbar}}~{\bf 11},  23765 (123).

\bibitem{loshchilov_decoupled_2019}
Loshchilov, I. and Hutter, F., ``Decoupled {Weight} {Decay} {Regularization},''
  (Jan. 2019).
\newblock arXiv:1711.05101 [cs, math].

\bibitem{shin_abnormal_2018}
Shin, Y., Qadir, H.~A., and Balasingham, I., ``Abnormal {Colon} {Polyp} {Image}
  {Synthesis} {Using} {Conditional} {Adversarial} {Networks} for {Improved}
  {Detection} {Performance},'' {\em IEEE Access}~{\bf 6},  56007--56017 (2018).

\end{thebibliography}
\bibliographystyle{spiebib} 

\end{document}